%% file: main.tex
\documentclass{article}

\usepackage[dvipsnames]{xcolor}

\PassOptionsToPackage{hyperfootnotes=false}{hyperref}
\usepackage[final]{corl_2026}
\usepackage{amsmath,amssymb}
\usepackage{xspace}

\usepackage{colortbl}
\usepackage{multirow}
\usepackage{mathtools}

\usepackage{wrapfig}

\usepackage{booktabs}
\usepackage{graphicx}
\usepackage{caption}
\usepackage[normalem]{ulem}

\usepackage{pgfplots}
\pgfplotsset{compat=1.17}
\definecolor{cOurs}{HTML}{D55E00}
\definecolor{cDA}{HTML}{0072B2}
\definecolor{cDF}{HTML}{009E73}
\definecolor{cHG}{HTML}{CC79A7}
\definecolor{cAbl}{HTML}{F0A860}

\usepackage{cleveref}
\crefname{figure}{Fig.}{Figs.}
\crefname{equation}{Eq.}{Eqs.}
\crefname{section}{Sec.}{Secs.}
\crefname{table}{Tab.}{Tabs.}

\newcommand{\pipeline}{{DreamStream}\xspace}
\newcommand{\metric}{{FD$\pi$}\xspace}
\newcommand{\singlemetric}{{\mathrm{FD}\pi_k}\xspace}
\newcommand{\eqmetric}{\text{FD}\pi}
\newcommand{\benchmark}{{Navhard-CL}\xspace}

\title{\pipeline: Towards Policy-Oriented Generative Simulation for End-to-End Driving}

\author{
  Ziyang Leng$^1$\thanks{Equal contribution.}
  \And
  Sicheng Mo$^1$\footnotemark[1]
  \And
  Seth Z. Zhao$^1$
  \And
  Haoyuan Cai$^1$
  \AND
  Yu Zeng$^2$
  \And
  Rowan McAllister$^2$
  \And
  Bolei Zhou$^1$
  \AND
  \normalfont $^1$University of California, Los Angeles \quad \quad \quad \quad $^2$Toyota Research Institute \\
}

\begin{document}
\maketitle

\begin{abstract}
Faithfully evaluating end-to-end driving policies in simulation requires observations that are not merely photo-realistic, but preserve the scene features a policy relies on to make decisions. Existing platforms, however, exhibit a sim-to-real visual gap that corrupts policy perception, undermining their ability to assess a policy's closed-loop decision-making. To this end, we propose \textbf{\pipeline}, a generative, closed-loop simulator that achieves \emph{policy-oriented fidelity} using a simulator-grounded autoregressive video model. Our video model is distilled from a large pretrained video model via traffic layout guidance, varying visual appearance while preserving policy-relevant features such as scenario layout and the temporal consistency of dynamic objects. We further observe that perceptual metrics like FID misrank how well these features are preserved. To tackle this, we introduce {\boldmath\textbf{\metric}}, a new multi-representation metric that measures the sim-to-real gap as the Fr\'echet distance over scene-context features from public E2E policies. Under \metric, \pipeline improves over the strongest prior closed-loop simulator by $1.6\times$ on nuScenes and $4.7\times$ on NAVSIM, and induces the least perturbation to policy's perceptual observability. Based on \pipeline, we construct \textbf{\benchmark} benchmark, which turns non-reactive real-world benchmark NAVSIM into interactive testing environments with adversarial driving behaviors and weather variations. This benchmark exposes many failure modes of driving policies, such as scorer bias and lack of recovery behaviors, that prior closed-loop benchmarks overlook. Code and data are available at \url{https://github.com/VAIL-UCLA/DreamStream}.

\end{abstract}


\input{tex/main/01_intro}
\input{tex/main/02_related_work}
\input{tex/main/03_approach}
\input{tex/main/04_experiments}
\input{tex/main/05_limitations}
\input{tex/main/06_conclusion}

\acknowledgments{
This work was supported by NSF grants CNS-2235012 and IIS-2339769, and Toyota Research Institute. Seth Z. Zhao was supported by Qualcomm Innovation Fellowship. Sicheng Mo was supported by Amazon AI PhD Fellowship through the Science Hub for Humanity and Artificial Intelligence.
}

\bibliography{references}
\newpage
\appendix
\input{tex/supp/appendix}

\end{document}

%% file: tex/main/01_intro.tex
\section{Introduction}
\label{sec:main-intro}

Recent studies~\cite{ol_cl_survey,wang2026open,zhao2026bridgesim} reveal a substantial open-loop (OL) to closed-loop (CL) evaluation gap, making closed-loop simulation necessary for the reliable assessment of end-to-end (E2E) driving policies. Since E2E policies map sensor observations directly to actions, a closed-loop simulator must provide controllable traffic scenarios, namely physically grounded layouts and agent behaviors, while rendering observations that are visually realistic and preserve the scene features a policy relies on to make decisions. Controllability of traffic scenarios is already well supported by physics-based simulators~\cite{li2022metadrive,kazemkhani2024gpudrive}, yet these leave a substantial sim-to-real visual appearance gap~\cite{jia2024bench,gerstenecker2026fail2drive,zhao2026bridgesim}; conversely, generative and reconstruction-based simulators improve realism but operate within narrow visual domains and often forfeit controllability~\cite{zhou2025hugsim,ni2025recondreamer,zhao2025drivedreamer4d,yang2025drivearena,mei2024dreamforge}. No existing platform delivers both, leaving closed-loop evaluation unable to thoroughly assess a policy's driving performance.

\begin{figure}[t]
\centering
\includegraphics[width=0.95\linewidth]{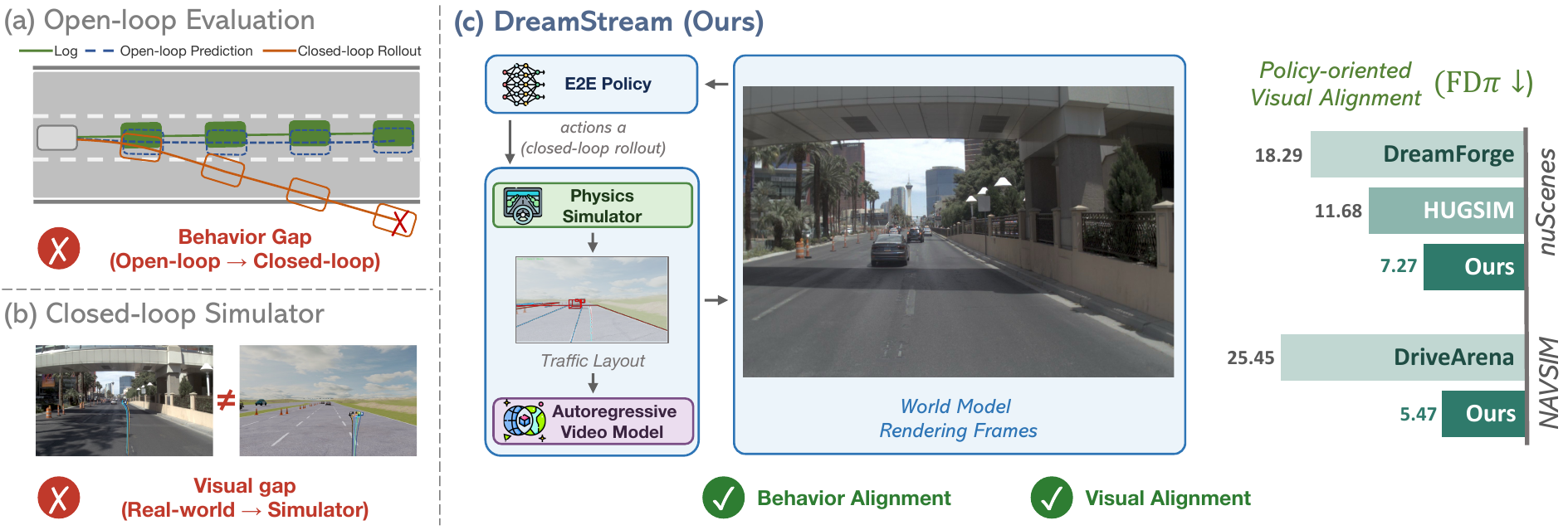}
\caption{\textbf{Motivation:} existing benchmarks for E2E driving either omit behavioral testing or evaluate unfaithfully due to the sim-to-real gap. \textbf{\pipeline} bridges sim-to-real via policy-oriented visual alignment and grounded traffic for faithful closed-loop evaluation.}
\label{fig:teaser}
\vspace{-2em}
\end{figure}

In this work, we present \textbf{\pipeline}, a generative closed-loop simulator that bridges the aforementioned evaluation gaps as illustrated in~\cref{fig:teaser}. \pipeline couples a physics simulator with an autoregressive video model with the following protocol: the physics simulator constructs and rolls out scenarios with interactive agents, and the autoregressive video model translates the simulator's symbolic state into photorealistic visual observations. Such a design ensures visual realism for ego-agent observation, while background traffic is logged using real-world dynamics.

Building \pipeline's video model imposes three requirements that previous generative simulators cannot meet simultaneously. \emph{(i) Autoregressive long-horizon rollout.} Closed-loop evaluation requires generating hundreds of consecutive rollout frames. We adopt an autoregressive design with early-timestep latent states as the KV cache to maintain consistency across the entire scenario. \emph{(ii) Scene layout alignment.} E2E policies attend to visual cues, such as lane geometry and traffic-element state, for decisions.
Without explicit conditioning on the simulator's layout, the video model may produce misaligned lane geometry and agent positions.
We therefore introduce \emph{traffic layout guidance} during distillation, applying classifier-free guidance on the layout condition so that rendered frames preserve simulator state. \emph{(iii) Diverse visual appearance.} Policies must be stress-tested under diverse and scarce visual conditions, but a video model trained only on small in-domain datasets~\cite{yang2025drivearena,mei2024dreamforge} inherits their narrow visual distribution. We propose \emph{diverse-scene distillation} to retain broad visual priors that are absent from the training data. Together, these designs enable \pipeline to render policy-decision-relevant, long-horizon, and visually diverse observations for closed-loop policy evaluation.

To verify a simulator achieves \emph{policy-oriented} visual alignment, we introduce \metric.
Prior simulators measure visual quality with FID~\cite{heusel2017fid} and controllability with 3D detection or map segmentation accuracy~\cite{gao2024magbeic_drive_v1,yang2025drivearena,mei2024dreamforge}, and recent benchmarks expand to multiple aspects~\cite{liang2025worldlens}. However, FID computed on small in-domain driving datasets is biased~\cite{mo2025dreamland,liang2025worldlens,yang2026fd-loss},
and controllability metrics reflect limited alignment dimensions, which may not be relevant to the policy's decision at all.
\metric instead measures visual alignment using the scene context feature from E2E policies, computing the Fr\'echet distance between this feature for paired real and simulated scenes.
As shown in \cref{fig:metric-correlation}, \metric exhibits a substantially stronger correlation with the driving performance than FID and FVD, confirming that it captures visual cues policies depend on.
Under \metric, \pipeline outperforms the strongest prior closed-loop simulator by $1.6\times$ on nuScenes and $4.7\times$ on NAVSIM.

Based on the diversity of visual appearance and the scenario customizability of \pipeline, we construct \textbf{\benchmark}, a closed-loop benchmark that \emph{turns the non-reactive, static open-loop benchmark NAVSIM~\cite{navsimv1} into reactive, dynamic closed-loop scenario testing environments}. \benchmark could be used to systematically diagnose the observation and behavior gaps that current E2E policies face under closed-loop deployment. It consists of a real-world log-replay base set \emph{Navhard-Base}~\cite{navsimv2} and two challenging variations: \emph{Navhard-AdvBehavior} introduces adversarial agents to manifest safety-critical interactions, and \emph{Navhard-AdvWeather} renders the same scenarios under diverse weather, lighting, and road-surface conditions. The benchmark reveals three failure modes under closed-loop settings: scorer miscalibration, proposal-coverage failure, and vision encoder brittleness under appearance shifts. These findings suggest that \benchmark could complement existing benchmarks for evaluating driving performance.
We summarize our contributions as follows:
\begin{itemize}
    \item  We design \textbf{\pipeline}, a generative closed-loop simulator that pairs a physics simulator with an autoregressive video model. \pipeline delivers both \emph{visual realism} and \emph{scenario realism} necessary for reliable closed-loop simulation for E2E driving policies.
    \item  We propose {\boldmath\textbf{\metric}}, a policy-oriented visual alignment metric for driving world models. Under \metric, \pipeline outperforms the strongest prior simulator by $1.6\times$ on nuScenes and $4.7\times$ on NAVSIM.
    \item We construct a closed-loop benchmark, \textbf{\benchmark}, built on \pipeline, with behavioral and visual variations. It reveals performance gaps and failure modes of existing driving policies that prior closed-loop benchmarks overlook.
\end{itemize}

%% file: tex/main/02_related_work.tex
\section{Related Work}
\label{sec:main-related}

\textbf{Autoregressive Video Generation for Autonomous Driving}.
Generating realistic driving observations for interactive evaluation remains an open problem, as closed-loop simulators must render action-conditioned camera views autoregressively while preserving real-world appearance and temporal coherence.
Three lines of work address this by predicting future frames with generative world models~\cite{kim2021drivegan,hu2023gaia,gao2024vista,gao2024magbeic_drive_v1,gao2024magicdrive_v2,wen2024panacea,lu2024infinicube,zheng2024genad}, synthesizing novel views from reconstructed dynamic 3D scenes~\cite{yang2024emernerf,zhou2025hugsim,ni2025recondreamer,zhao2025drivedreamer4d}, and re-rendering physics-based simulator state with large pretrained video models~\cite{zhou2024simgen,mo2025dreamland,mei2024dreamforge,yang2025drivearena}.
However, existing pipelines are typically trained on small driving datasets such as nuScenes and inherit their narrow visual style, failing to support diverse weather, lighting, and road conditions during closed-loop evaluation~\cite{caesar2020nuscenes,yang2025drivearena,mei2024dreamforge}.
\pipeline mitigates this with diverse-scene distillation in a multi-stage autoregressive pipeline distilled from a large pretrained video foundation model~\cite{wan2025wan}, enabling weather and lighting variations for evaluations.

\textbf{End-to-End Driving Policies Evaluation}. Open-loop evaluation emerged as a practical proxy for closed-loop simulation, enabling efficient benchmarking of planning policies on large-scale datasets~\cite{caesar2020nuscenes, caesar2021nuplan, navsimv1, navsimv2, li2024pretrain, zhao2025quantv2x}, where interactive evaluation is computationally expensive and difficult to standardize~\cite{dolgov2008practical, claussmann2019review}. However, because it evaluates policies under fixed, non-reactive observations rather than sequential decision-making, strong open-loop performance may not be indicative of reliable driving behavior in closed-loop deployment~\cite{jia2024bench, li2024ego, ol_cl_survey, zhao2026bridgesim, coscoy2026mdrive}. This motivates closed-loop evaluations of current E2E policies that are widely evaluated in an open-loop manner~\cite{RAP, DrivoR, zhou2025autovla, navsimv1, diffusiondrive}. To this end, \pipeline provides a generative simulator that bridges the sim-to-real visual gap while grounded in closed-loop simulation dynamics, thus enabling fair and comprehensive closed-loop evaluations of E2E driving policies.

%% file: tex/main/03_approach.tex
\section{\pipeline Approach}
\label{sec:main-approach}

We give an overview of our closed-loop generative simulator \pipeline in \cref{sec:main-approach-overview},
and then describe the multi-stage distillation pipeline of the autoregressive video diffusion model in \cref{sec:main-approach-distillation}.

\begin{figure*}[t]
    \centering
    \includegraphics[width=0.97\linewidth]{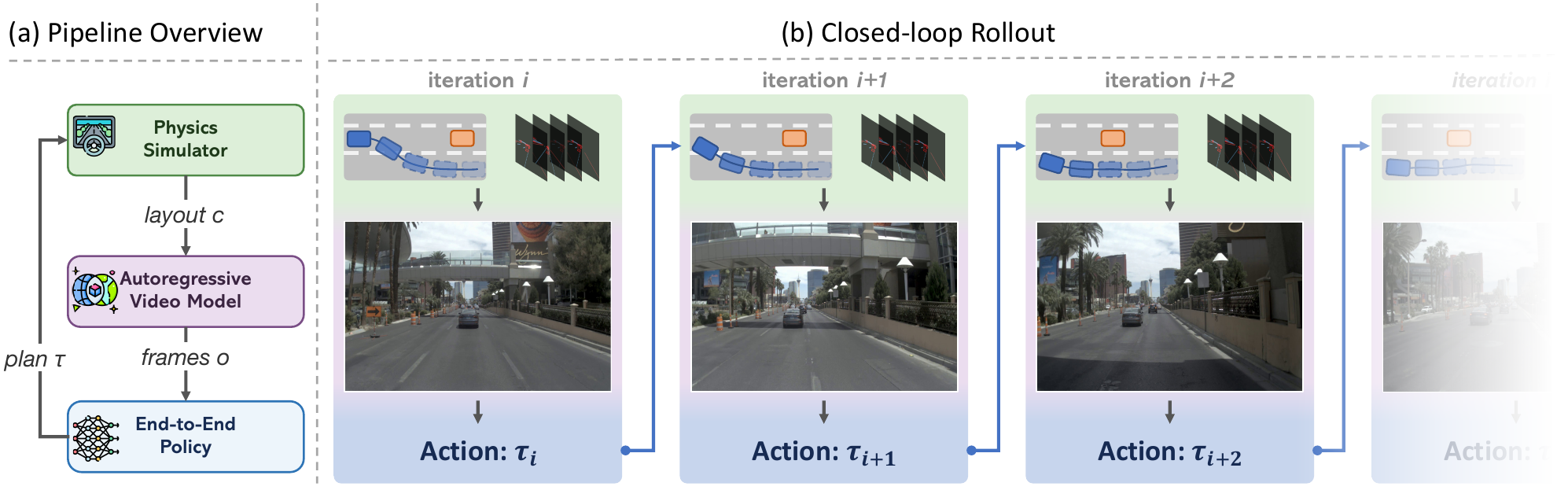}
    \caption{
   Overview of the \pipeline framework. At each iteration of closed-loop simulation, it performs autoregressive rollouts with layout from a physics simulator and diverse visual appearance.
   }
   \label{fig:pipeline}
   \vspace{-1em}
\end{figure*}

\subsection{Overview}
\label{sec:main-approach-overview}

\pipeline consists of closed-loop interaction between three components: a driving simulator $\mathcal{S}$, our autoregressive (AR) video model $\mathcal{G}_\phi$ (\cref{sec:main-approach-distillation}), and an E2E driving policy $\pi$ given for evaluation. The simulator maintains the underlying scene state, including ego pose, surrounding agents, and an HD map.
The video model translates the simulator state into the camera frames that the policy actually consumes. This decoupling lets \pipeline inherit the simulator grounding while providing high visual fidelity.
\cref{fig:pipeline} illustrates the iteration cycle.

\textbf{Scenario initialization.} A rollout begins from a scenario specified by an initial world state $\mathbf{s}_0$ together with an initial camera frame $\mathbf{o}_0$. We support two scenario sources: (i) real-world driving logs converted into the simulator~\cite{li2023scenarionet}, where $\mathbf{o}_0$ is the corresponding logged frame, and (ii) generated scenarios such as safety-critical variants~\cite{advbmt}, where $\mathbf{o}_0$ is synthesized by an image generation model to match the desired initial conditions.

\textbf{Closed-loop iteration.} Each iteration $i$ starts from time $t_i$ with the world state $\mathbf{s}_{t_i}$ and the most recent $K$ camera frames $\mathbf{o}_{t_i - K + 1 : t_i}$, and proceeds in three steps.

\noindent\emph{(1) Plan.} The E2E policy produces a planned trajectory of length $H$ steps, $\tau_{t_i} = \pi\!\left( \mathbf{o}_{t_i - K + 1 : t_i} \right) = \left( \mathbf{a}_{t_i+1}, \ldots, \mathbf{a}_{t_i + H} \right)$.

\noindent\emph{(2) Simulator rollout.} The simulator executes only the first $\Delta_r \leq H$ steps of $\tau_{t_i}$, where $\Delta_r$ is a configurable \emph{replan period} that controls how often the policy is re-queried: $\mathbf{s}_{t_i + k} = \mathcal{S}\!\left( \mathbf{s}_{t_i + k - 1}, \mathbf{a}_{t_i + k} \right), k = 1, \ldots, \Delta_r$.
For each new state, the simulator renders a traffic-layout condition $\mathbf{c}_{t_i + k}$, which contains the perspective projection of the high-definition (HD) map (lane geometry and traffic elements) together with 3D bounding boxes of traffic agents (vehicles and pedestrians).

\noindent\emph{(3) Observation generation.} The video model autoregressively generates the next $\Delta_r$ camera frames conditioned on the rendered layouts and a text prompt $\mathbf{p}$ describing the scene: $\mathbf{o}_{t_i + 1 : t_i + \Delta_r} = \mathcal{G}_\phi\!\left( \mathbf{c}_{t_i + 1 : t_i + \Delta_r}, \mathbf{p} \;\big|\; \mathcal{K}_i \right)$,
where $\mathcal{K}_i$ is the video model's KV cache that grows with each iteration's generated frames. $\mathcal{K}_0$ is seeded with the initial frame $\mathbf{o}_0$.
The policy then takes the most recent $K$ camera frames as its next input,
and the loop continues from $t_{i+1} = t_i + \Delta_r$.

\textbf{Closed-loop scoring.} During rollout, surrounding agents are controlled via log-replay, intelligent driver model (IDM), or adversarial modes. After the scenario terminates, the full executed trajectory $\{\mathbf{s}_t\}$ is scored with closed-loop metrics.

\subsection{Autoregressive Video Diffusion Model Distillation}
\label{sec:main-approach-distillation}

As training a few-step autoregressive video diffusion model from scratch is difficult, we follow common practice and adopt a three-stage distillation recipe~\cite{huang2026self,shin2025motionstream,gao2026dreamdojo}: Stage-1 builds a controllable conditional video generator; Stage-2 turns it into a few-step autoregressive model for efficient rollout; Stage-3 finally improves its long-horizon rollout stability.

Starting from a pretrained video model, we define the input as \(c=(\mathbf{c},\mathbf{o}_0)\), where \(\mathbf{c}\) denotes traffic-layout conditions and \(\mathbf{o}_0\) is the first-frame anchor, and train \(G_{\mathrm{T}}\) to generate future video \(\hat{\mathbf{o}}_{1:F}\) that follows \(c\) while preserving realistic appearance and dynamics. Stage-1 teaches the model to generate the future video from explicit conditions, so controllability is learned before causal distillation.

Stage-2 and Stage-3 share the same goal: obtaining a stable few-step causal AR model \(G_\theta\) for long rollout. Stage-2 performs teacher-to-student distillation with ground-truth context. It uses \((\mathbf{x}^{\mathrm{T}}_t, c, t)\) as input and \(\mathbf{x}_0\) as target, where \(c=(\mathbf{c},\mathbf{o}_0)\), which gives the causal student a strong initialization for chunk-level AR prediction while inheriting the teacher's generation quality. Stage-3 then addresses the remaining train-test gap by training on self-generated context (Self Forcing) rather than ground-truth history. In this stage, the rollout is conditioned on \(c\), the AR clean output is \(\hat{\mathbf{x}}_0^{1:F}=\mathcal{R}_\theta(c)\), and training is applied on its noised version \(\hat{\mathbf{x}}_t\), so the model is explicitly optimized under its own rollout distribution. Detailed formulations are provided in the Appendix \cref{sec:supp-distillation-details}.

Alongside the multi-stage distillation pipeline, we further introduce two key components to enhance the distilled autoregressive video diffusion model to preserve policy-oriented fidelity.

\textbf{Traffic-guided distillation}.
In closed-loop simulation, the policy plans based on world-model-generated camera frames, so misalignment in lanes or agents can change decisions even when the simulator state is correct.
The distilled model must therefore reliably follow the traffic layout in $c$.
Following classifier-free guidance~\cite{ho2022classifier}, extrapolating between conditional and unconditional estimates steers denoising toward greater satisfaction of the conditioning. Thus, it mimics a classifier gradient without training a separate classifier.
During Stage-1 training, we replace the layout in $c$ with $c_{\emptyset}$ with probability $p{=}0.1$ where $c_{\emptyset}$ removes traffic layout but keeps $\text{CLIP}(\mathbf{o}_0)$.
When generating videos with $G_{\mathrm{T}}$, we combine predictions under $c$ and $c_{\emptyset}$ with guidance scale $w$,
\begin{equation}
    \hat{G}_{\mathrm{T}}(\mathbf{x}_t, c, t) = G_{\mathrm{T}}(\mathbf{x}_t, c_{\emptyset}, t) + w \cdot \big( G_{\mathrm{T}}(\mathbf{x}_t, c, t) - G_{\mathrm{T}}(\mathbf{x}_t, c_{\emptyset}, t) \big).
\end{equation}
Without additional data, this guidance improves traffic layout alignment in $G_{\mathrm{T}}$ distillation pipeline.

\textbf{Diverse-scene distillation}.
Pretrained video models already learned to generate diverse weather and lighting conditions.
However, our distillation data comes from driving logs with limited coverage, and most training scenes share similar weather and lighting.
When we distill the teacher into the causal student model, the autoregressive student model gradually loses the ability to render adverse or visually diverse conditions that were present in the pretrained backbone but absent from the driving dataset.

To obtain visually diverse distillation data without new driving logs, we build synthetic clips from existing driving scenarios detailed in \cref{sec:supp-synthetic-clips}.
Training Stage-2 and 3 on these clips enables \pipeline to roll out under conditions outside the narrow visual domain of small-scale driving data.

%% file: tex/main/04_experiments.tex
\section{Policy-oriented World Model Evaluation}
\label{sec:main-wm-eval}

Existing world-model evaluation metrics fail to capture whether a model preserves the information downstream policies rely on for decisions. Perceptual metrics such as FID~\cite{heusel2017fid} were designed for visual quality rather than whether the generated world supports downstream autonomy. They saturate against a single feature space and are biased on small in-domain driving datasets~\cite{mo2025dreamland,liang2025worldlens,yang2026fd-loss}. Controllability evaluations, such as 3D detection or map segmentation accuracy~\cite{gao2024magbeic_drive_v1,yang2025drivearena,mei2024dreamforge, liang2025worldlens}, measure task-specific alignment from a modular rather than an E2E perspective. Therefore, we need a direct and unified visual alignment metric from the E2E policies' perspective.

\subsection{Metric Design}
\label{sec:main-wm-eval-metric}

We design \metric that quantifies the sim-to-real visual gap for closed-loop policy evaluation from the policy's perspective.
For a driving policy, we extract the scene-context features it uses to generate actions from real camera frames and world-model-rendered frames of the same scenes, and compute the Fr\'echet distance between the resulting feature distributions. This captures how much the world model perturbs the visual information the policy uses to act. Different E2E policies attend to various aspects (appearance, geometry, critical objects, traffic semantics) of a driving scene differently; we therefore aggregate the measurement across a panel of public E2E policies~\cite{DrivoR, diffusiondrive, transfuser, RAP, sun2026sparsedrivev2}.

\textbf{Formulation.}
Let $\Pi = \{\pi_1, \ldots, \pi_N\}$ denote a panel of E2E driving policies, $\mathcal{T}$ a validation scene token set, and $v$ the world model under evaluation. For each token $t \in \mathcal{T}$, we extract the scene context feature of the original camera frame as $\Phi_{\pi_k}(t) \in \mathbb{R}^{D_{\pi_k}}$, with $D_{\pi_k}$ policy-dependent. We also extract  scene context feature $\tilde{\Phi}_{\pi_k}^{(v)}(t)$ of the generated frame. Modeling $\Phi_{\pi_k}$ and $\tilde{\Phi}_{\pi_k}^{(v)}$ as samples from multivariate Gaussians with moments $(\mu_{\pi_k}, \Sigma_{\pi_k})$ and $(\tilde{\mu}_{\pi_k}^{(v)}, \tilde{\Sigma}_{\pi_k}^{(v)})$, we compute the Fr\'echet distance
\begin{equation}
    \widetilde{\mathrm{FD}}{\pi_k}{(v)} \;=\; \lVert \mu_{\pi_k} - \tilde{\mu}_{\pi_k}^{(v)} \rVert_2^2 \;+\; \mathrm{Tr}\!\left( \Sigma_{\pi_k} + \tilde{\Sigma}_{\pi_k}^{(v)} - 2\bigl(\Sigma_{\pi_k} \tilde{\Sigma}_{\pi_k}^{(v)}\bigr)^{1/2} \right).
\end{equation}
This measures the shift in the visual representation that policy ${\pi_k}$ actually consumes. The raw $\widetilde{\mathrm{FD}}{\pi_k}{(v)}$ scales with the feature norm of policy ${\pi_k}$, which varies across model architectures. To make it comparable across policies, we normalize it with $\mathrm{FD}{\pi_k}{(v)} = \widetilde{\mathrm{FD}}{\pi_k}{(v)} / \mathrm{Tr}(\Sigma_{\pi_k})$,
and report \metric by averaging across the policy panel:
\begin{equation}
    \eqmetric(v) \;=\; \frac{1}{|\Pi|} \sum_{{\pi_k} \in \Pi} \mathrm{FD}{\pi_k}{(v)}.
\end{equation}
Lower \metric indicates that the generated frames preserve more of the visual information that downstream policies rely on,
with $0$ marking perfect alignment between real and rendered representations.
Following existing distributional distance metrics~\cite{binkowski2018demystifying,jayasumana2024rethinking}, we report \metric $\times10^2$ for readability.
As demonstrated in~\cref{fig:metric-correlation}, \metric correlates substantially more strongly with driving performance than FID and FVD, suggesting that it better captures the visual cues relevant to policy behavior.

\input{tex/tables/fd_main}

\begin{figure*}[!t]
    \centering
    \includegraphics[width=0.99\linewidth]{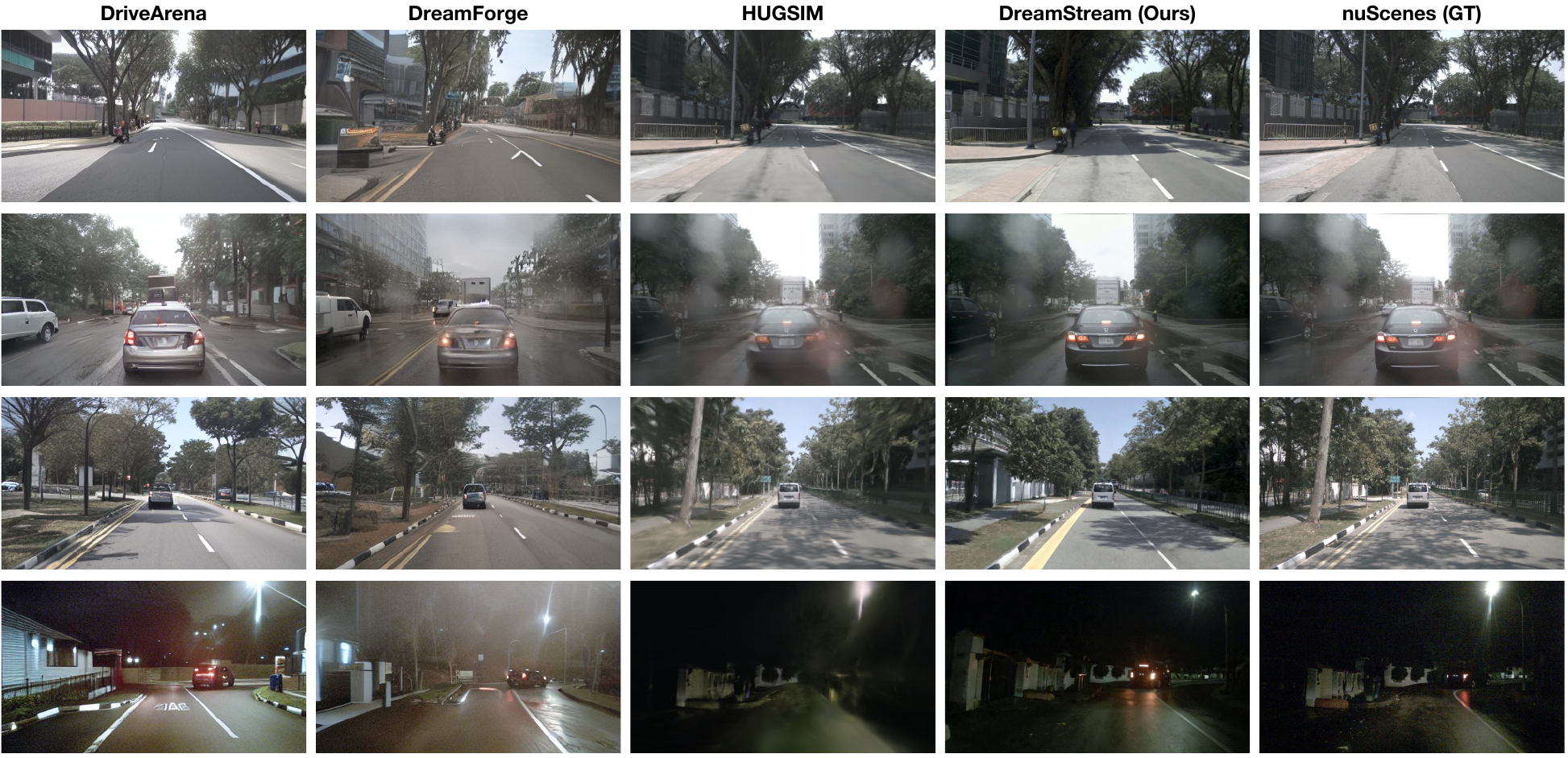}
    \caption{
   Qualitative comparison of our \pipeline and baselines on nuScenes \emph{val} scenarios.
   }
   \label{fig:qual-comparison}
    \vspace{-1em}
\end{figure*}

\subsection{Results}
\label{sec:main-wm-eval-results}

We compare our \pipeline against prior works on nuScenes \emph{val}~\cite{caesar2020nuscenes} and NAVSIM \emph{navtest}~\cite{navsimv1}, including MagicDrive~\cite{gao2024magbeic_drive_v1}, Panacea~\cite{wen2024panacea}, Dreamland~\cite{mo2025dreamland}, DriveArena~\cite{yang2025drivearena}, DreamForge~\cite{mei2024dreamforge}, and HUGSIM~\cite{zhou2025hugsim}. The policy panel comprises five distinct E2E architectures: DrivoR~\cite{DrivoR}, DiffusionDrive/DiffusionDriveV2 (DD/DDv2)~\cite{diffusiondrive,diffusiondrivev2}, LTF~\cite{transfuser}, RAP~\cite{RAP}, and SparseDriveV2 (SDv2)~\cite{sun2026sparsedrivev2}. Results are in~\cref{tab:fd-main}; FID and FVD are reported alongside for reference.

On nuScenes \emph{val}, \pipeline achieves $\eqmetric$ of 7.27, which is $1.6\times$ lower than the strongest baseline. The advantage holds on three of the five per-policy columns, with the largest reduction on DiffusionDrive. On RAP, reconstruction-based HUGSIM performs on par with ours, which demonstrates our world model's ability to preserve the 3D geometry that RAP's spatial cross-attention relies on.
The averaged \metric across diverse E2E policies unifies different architectures, and our model consistently outperforms previous baselines.
On NAVSIM \emph{navtest}, our model achieves the highest performance consistently. DriveArena's $\eqmetric$ increases $1.6\times$ than its nuScenes value, indicating its wider visual gap as the evaluation domain spans. Qualitative comparison in \cref{fig:qual-comparison} further confirms the strong visual alignment of \pipeline compared with baseline methods.

\subsection{Metric Ablation}

\textbf{Correlation.}
To further verify \metric{} correlation, we compute correlation in a held-out manner: for each policy $\pi$, we correlate the FD computed \emph{excluding} $\pi$ with $\pi$'s PDMS, so the features and the behavior come from different networks.
Held-out \metric remains strongly correlated ($r=-0.65$), whereas FID averages $r=+0.11$ and FVD $r=+0.08$, indicating that \metric captures policy-independent corruption of scene information.

\textbf{Sensitivity.}
We evaluate the sensitivity of \metric to challenging scenes and safety-critical local errors. When applying it to Navhard subsets of \emph{navtest} with 540 challenging scenes, \metric increases for 20\% more compared with a randomly sampled subset, which corresponds to the larger visual gap. FID on Navhard subsets only captures average representation shift.
We further inject local corruptions using~\cite{zhao2026precise} on $1{,}517$ nuScenes \emph{val} samples, which includes agent displacement for 2m and lane removal. Measuring on the corrupted samples, \metric responds $2.3\times$ more strongly than FID, showing higher sensitivity to policy-relevant local errors.

\begin{figure}[t]
\centering
\includegraphics[width=\linewidth]{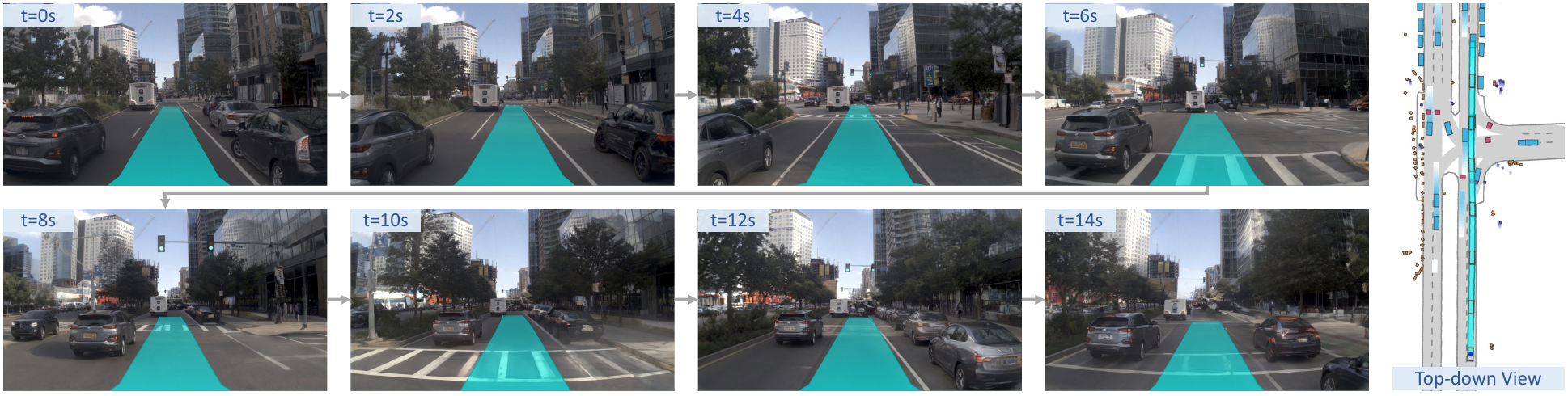}
\vspace{-1.5em}
\caption{Qualitative example of closed-loop policy evaluation using \pipeline.
}
\label{fig:cl-rollout}
\vspace{-1.5em}
\end{figure}

\section{\benchmark Closed-loop Benchmark}
\label{sec:main-benchmark}

Built on our \pipeline generative simulator, we create \benchmark, a closed-loop benchmark for evaluating end-to-end driving policies with realistic and diverse traffic scenarios and visual appearance. \benchmark comprises three scenario buckets based on NAVSIM \emph{navhard}~\cite{navsimv2}: \emph{Navhard-Base} from real-world logs, \emph{Navhard-AdvBehavior} which introduces adversarial agents to manifest safety-critical interactions, and \emph{Navhard-AdvWeather} that exposes policies to adverse and scarce weather, lighting, and road-surface conditions. We detailed scenario curation in Appendix \cref{sec:supp-benchmark-scenarios}.

\input{tex/tables/cl_gap_analysis}
\subsection{E2E Policies Performance}
\label{sec:main-benchmark-perf}

We evaluate E2E driving policies on the \emph{Navhard-Base} scenarios of \benchmark, comparing three observation sources: simulator RGB rendering from BridgeSim~\cite{zhao2026bridgesim}, DriveArena-rendered~\cite{yang2025drivearena}, and \pipeline-rendered in~\cref{fig:cl-rollout}. \cref{tab:cl-gap-analysis}a reports the closed-loop performance~\cite{navsimv2,zhao2026bridgesim}, including driving score (DS), extended PDM score (EPDMS), and route completion (RC). We find that better visual alignment in \pipeline yields higher closed-loop scores across evaluated policies, indicating that preserving \emph{policy-oriented visual fidelity} reduces simulation-induced perturbation to policy behavior and is essential for faithful closed-loop evaluation. Notably, policies score lower under the more photorealistic DriveArena than under the game-engine-rendered BridgeSim, suggesting that photorealism without policy-relevant scene-context preservation can degrade evaluation faithfulness even relative to a simulator with a visible sim-to-real gap.

\subsection{Closed-loop Gap Analysis}
\label{sec:main-benchmark-gap}

\benchmark reveals closed-loop performance gap and failure modes that prior benchmarks overlook. \emph{Navhard-AdvBehavior} decomposes the performance drop into two distinct observations by analyzing the policy's scorer and trajectory. \emph{Navhard-AdvWeather} measures policy brittleness under appearance shifts, such as adverse weather and on-road conditions. We refer to \cref{sec:supp-gap-analysis} for detailed setup, results, and analysis.

\textbf{Scorer bias and proposal coverage failure (\cref{tab:cl-gap-analysis}b).} We replace the policy's learned scorer with an oracle scorer that ranks proposals by ground-truth EPDMS calculated in the simulator. The driving score difference reveals two gaps: \emph{(i) Scorer bias} is already large on \emph{Navhard-Base}, and widens under \emph{Navhard-AdvBehavior} for certain architectures. \emph{(ii) Proposal coverage} fails even when scoring is optimal. The driving score decreases dramatically from \emph{Navhard-Base} to \emph{Navhard-AdvBehavior}. Decomposing by subscores shows lower TTC, LK, and HC, indicating the policy's inability to react safely.
LTF, a policy with no scoring head, shows that the gap lies in the trajectory decoder. These extend the open-loop scorer-mismatch analysis~\cite{ang2026clover} into the closed-loop regime.

\textbf{Visual robustness gap under appearance shift (\cref{tab:cl-gap-analysis}c).} For high driving score ($\geq 80$) scenarios in \emph{Navhard-AdvWeather}, rain and snow degrade DrivoR's performance more than night, especially on road-surface related performance (drivable-area compliance, lane keeping). It reveals a visual robustness gap that closed-loop benchmarks on a narrow distribution-matched visual domain cannot expose.

\section{Ablations}
\label{sec:main-ablation}

We ablate our key design for enhancing the distilled autoregressive video model, including the traffic-layout guidance and diverse-scene distillation using the Wan~2.1 backbone.

\begin{figure}[t]
\vspace{-0.5em}
\centering
\includegraphics[width=\linewidth]{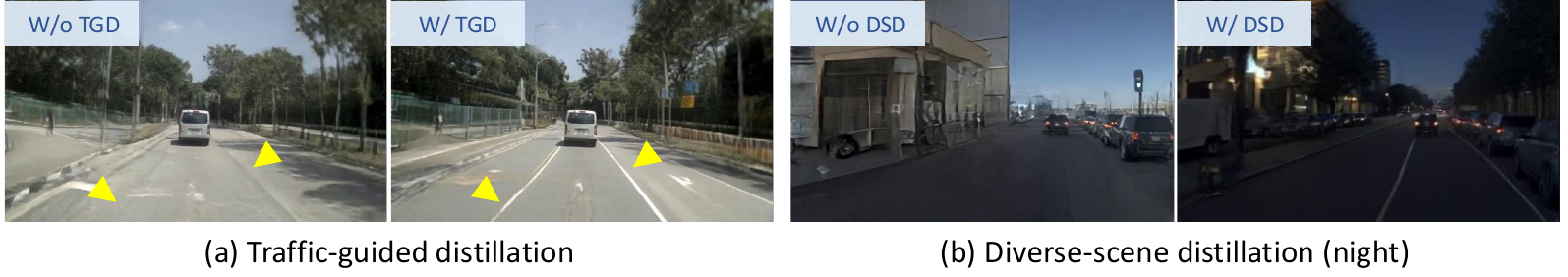}
\vspace{-1.5em}
\caption{Qualitative ablation on traffic-layout guidance and diverse-scene distillation.
}
\label{fig:qual-ablation}
\end{figure}

\label{sec:main-ablation-guidance}
\noindent\textbf{Traffic-guided distillation (TGD)} enhances the traffic layout signal in the generated frames, as shown on the left of \cref{fig:qual-ablation}.
This results in better scene layout alignment between the generated frames and simulator states, which benefits the evaluated E2E policies for more accurate decisions.
We ablate this design on nuScenes \emph{val} dataset and observe $\eqmetric$ improves from 12.18 to 9.50.

\label{sec:main-ablation-diverse-scene}
\textbf{Diverse-scene distillation (DSD)} is crucial for preserving visual diversity in generated frames. We ablate this design on \emph{Navhard} by conditioning on an edited first frame, and report the CLIP similarity between the generated frame at rollout step $100$ and the weather/lighting text prompt. DSD consistently improves CLIP similarity across diverse weather and lighting conditions. Without DSD, the autoregressive model fails to preserve the weather and lighting in the first frame and quickly reverts to the narrow appearance of the training distribution, as shown on the right of \cref{fig:qual-ablation}.

\input{tex/tables/ablation}

\begin{figure}[!t]
  \centering
  \vspace{-1em}
  \begin{minipage}[c]{0.64\linewidth}
    \centering
    \resizebox{\linewidth}{!}{
    \begin{tikzpicture}
      \begin{axis}[
      width=8.6cm, height=4.2cm,
      xlabel={Rollout Frames},
      xlabel near ticks, xlabel shift=-3pt,
      ylabel={\metric{}},
      ylabel style={
        rotate=-90,
        at={(axis description cs:-0.1,0.9)},
        anchor=south west,
        yshift=1pt, font=\scriptsize,
      },
      label style={font=\footnotesize},
      tick label style={font=\scriptsize},
      xmin=48, xmax=202,
      ymin=0, ymax=30,
      xtick={50,75,100,125,150,175,200},
      ytick={0,5,10,15,20,25},
      grid=major, grid style={line width=0.4pt, draw=gray!20},
      axis line style={gray!60},
      legend style={
        font=\fontsize{5}{5.5}\selectfont,
        draw=gray!40, fill=white, fill opacity=0.85, text opacity=1,
        at={(0.5,0.74)}, anchor=center,
        legend columns=-1,
        inner xsep=1pt, inner ysep=-0.5pt,
        /tikz/every even column/.append style={column sep=1.5pt},
      },
      legend image post style={mark size=1pt, line width=0.6pt},
      legend image code/.code={\draw[mark repeat=2, mark phase=2, #1]
        plot coordinates {(0cm,0cm) (0.07cm,0cm) (0.14cm,0cm)};},
      legend cell align=left,
      every axis plot/.append style={line width=0.9pt},
    ]
      \addplot[cOurs, line width=1.1pt, mark=*, mark size=1pt]
        coordinates {(50,6.04)(75,6.45)(100,6.64)(125,6.82)(150,6.95)(175,7.19)(200,7.27)};
      \addlegendentry{\pipeline}
      \addplot[cAbl, line width=1.1pt, mark=star, mark size=1.8pt]
        coordinates {(50,6.98)(75,7.84)(100,8.85)(125,9.61)(150,10.03)(175,10.57)(200,10.78)};
      \addlegendentry{\pipeline w/o cache}
      \addplot[cDA, mark=square*, mark size=1pt]
        coordinates {(50,16.77)(75,16.26)(100,16.06)(125,15.93)(150,15.79)(175,15.72)(200,15.68)};
      \addlegendentry{DriveArena}
      \addplot[cDF, mark=triangle*, mark size=1pt]
        coordinates {(50,19.12)(75,18.94)(100,18.76)(125,18.73)(150,18.54)(175,18.40)(200,18.29)};
      \addlegendentry{DreamForge}
      \addplot[cHG, mark=diamond*, mark size=1pt]
        coordinates {(50,13.87)(75,13.18)(100,12.77)(125,12.42)(150,12.10)(175,11.79)(200,11.68)};
      \addlegendentry{HUGSIM}
      \addplot[cOurs, line width=1.1pt, dashed, mark=o, mark size=1.4pt,
               mark options={solid, line width=0.9pt}]
        coordinates {(50,3.71)(100,4.34)(150,5.00)(200,5.47)};
      \addplot[cAbl, line width=1.1pt, dashed, mark=star, mark size=1.8pt,
               mark options={solid, line width=0.7pt}]
        coordinates {(50,4.16)(100,5.59)(150,6.77)(200,7.53)};
      \addplot[cDA, dashed, mark=square, mark size=1.4pt,
               mark options={solid, line width=0.8pt}]
        coordinates {(50,26.98)(100,26.03)(150,25.68)(200,25.45)};
      \node[anchor=south east, draw=gray!40, fill=white,
            fill opacity=0.85, text opacity=1,
            inner xsep=2pt, inner ysep=1pt,
            font=\fontsize{5}{5.5}\selectfont]
        at (rel axis cs:0.985,0.02)
        {\tikz[baseline=-0.55ex]{\draw[black,line width=0.6pt](0pt,0pt)--(5pt,0pt);}~Solid: nuScenes\quad
         \tikz[baseline=-0.55ex]{\draw[black,line width=0.6pt,dash pattern=on 1.6pt off 1.1pt](0pt,0pt)--(5pt,0pt);}~Dashed: NAVSIM};
    \end{axis}
    \end{tikzpicture}
    }
    \vspace{-2em}
    \caption{\metric{} over long-horizon rollout.\label{fig:rollout-drift}}
  \end{minipage}%
  \hfill
  \begin{minipage}[c]{0.28\linewidth}
    \centering
    \setlength{\tabcolsep}{2pt}
    \renewcommand{\arraystretch}{0.95}
    \begin{tabular}{@{}lrr@{}}
    \toprule
             & Wan2.1         & Wan2.2 \\
             & 1.3B           & 5B \\
    \midrule
    ms/frame & 49.5           & 59 \\
    Gen.\ fps & 20.2          & 16.9 \\
    VRAM     & 32 GB          & 67 GB \\
    Sim.\ fps  & 10.1         & 8.8 \\
    \bottomrule
  \end{tabular}
  \captionof{table}{\pipeline efficiency.}\label{tab:efficiency}
  \end{minipage}
  \vspace{-1.5em}
\end{figure}

\textbf{Long-horizon rollout stability.}
We quantify autoregressive quality drift by computing \metric as a function of rollout length in \cref{fig:rollout-drift}. \pipeline's drift is bounded and remains below all compared baselines on both nuScenes and NAVSIM. Using early-timestep latent states as the KV cache mitigates drift by $68\%$ on nuScenes and $48\%$ on NAVSIM for longer rollouts.

\textbf{Efficiency.}
\cref{tab:efficiency} reports the runtime of \pipeline on a single RTX~PRO~6000 GPU. Our design enables linear multi-GPU scaling, real-time generation, and batch benchmarking. Full model training costs ${\sim}200$ A100 GPU-days.

%% file: tex/tables/fd_main.tex
\begin{figure}[t]
\begin{minipage}[t]{0.62\linewidth}
  \vspace{2pt}
  \centering

  \small
  \setlength{\tabcolsep}{4.5pt}
  \resizebox{\linewidth}{!}{%
  \begin{tabular}{l|c|ccccc|cc}
    \toprule
 & & \multicolumn{5}{c|}{$\singlemetric$$\downarrow$} & \\
Method & \metric{}$\downarrow$ & DrivoR & DD & LTF & RAP & SDv2 & FID$\downarrow$ & FVD$\downarrow$ \\
\midrule
\rowcolor{gray!10} \multicolumn{9}{l}{nuScenes \emph{val}} \\
MagicDrive~\cite{gao2024magbeic_drive_v1}  & 17.18 & 11.76 & 31.74 & \uwave{4.57} & 32.00 & \textbf{5.84} & \underline{16.20} & 218.12 \\
Panacea~\cite{wen2024panacea}              & 31.46 & 19.68 & 57.44 & 8.47 & 54.81 & 16.89 & \uwave{16.96} & \textbf{139.00} \\
Dreamland~\cite{mo2025dreamland}           & 25.28 & 30.69 & 30.69 & 7.40 & \uwave{24.18} & 33.46 & 47.93 & 670.90 \\
DriveArena$^*$~\cite{yang2025drivearena}   & \uwave{15.68} & \underline{9.95} & \uwave{27.55} & 4.68 & 25.66 & 10.54 & 34.74 & 665.17 \\
DreamForge$^*$~\cite{mei2024dreamforge}    & 18.29 & \uwave{11.71} & 31.91 & \underline{4.09} & 36.70 & \underline{7.04} & \textbf{14.61} & \uwave{209.90} \\
HUGSIM$^*$~\cite{zhou2025hugsim}           & \underline{11.68} & 12.39 & \underline{23.76} & 5.79 & \textbf{8.25} & 8.21 & 27.95 & \underline{147.18} \\
\textbf{\pipeline} & \textbf{7.27} & \textbf{9.42} & \textbf{8.31} & \textbf{2.25} & \underline{9.24} & \uwave{7.12} & 19.58 & 272.22 \\
\midrule
\rowcolor{gray!10}\multicolumn{9}{l}{NAVSIM \emph{navtest}} \\
BridgeSim~\cite{zhao2026bridgesim} & 56.13 & 57.04 & 77.00 & 22.15 & 60.92 & 63.58 & 175.53 & - \\
DriveArena~\cite{yang2025drivearena}       & 25.45 & 25.18 & 38.96 & 10.17 & 30.78 & 22.16 & 41.80 & - \\
\textbf{\pipeline} & \textbf{5.47} & \textbf{5.21} & \textbf{9.69} & \textbf{2.65} & \textbf{5.05} & \textbf{4.74} & \textbf{11.78} & - \\
    \bottomrule
  \end{tabular}}
  \captionof{table}{\metric and $\singlemetric$ between original and generated/rendered frames.  \textbf{Bold}, \underline{underline}, and \uwave{wavy} marks first, second, third place, respectively.
  $^*$ trained on evaluation set.}
  \label{tab:fd-main}
\end{minipage}\hfill
\begin{minipage}[t]{0.36\linewidth}
  \vspace{0pt}
  \centering
  \includegraphics[width=\linewidth, trim=5 0 0 0, clip]{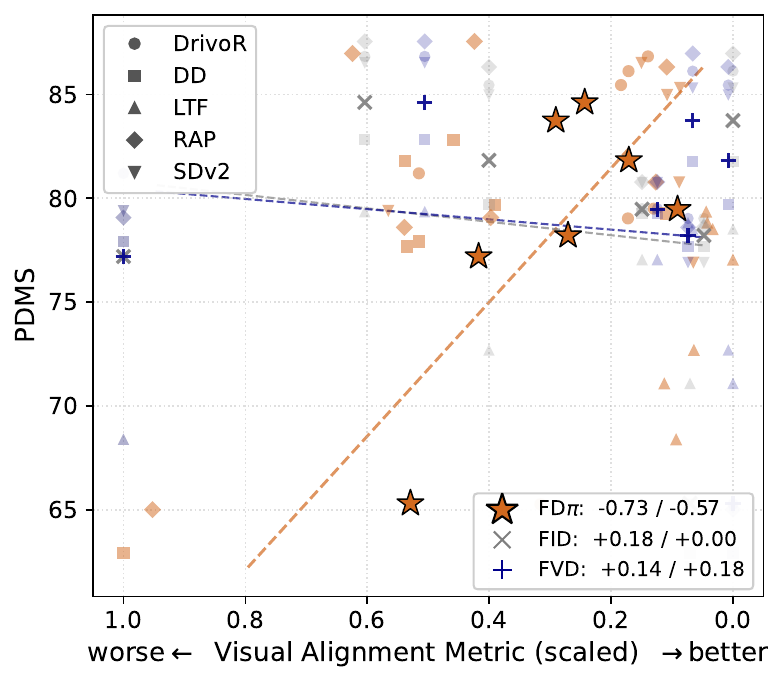}
  \caption{Correlation of \metric, FID, and FVD with OL PDMS.}
  \label{fig:metric-correlation}
\end{minipage}
\vspace{-1em}
\end{figure}

%% file: tex/tables/cl_gap_analysis.tex
\begin{table}[t]
\centering
\caption{Closed-loop performance and analysis on \benchmark. (a) E2E driving policy performance on \emph{Navhard-Base} under three observation sources. Sim: simulator RGB; DA: DriveArena; Ours: \pipeline. (b) Driving score (DS) gap due to scorer bias (Oracle$-$Learned) and proposal coverage (AdvBehavior$-$Base with oracle scorer) on \emph{Navhard-AdvBehavior}. $^{\dagger}$ has no scoring head.
}
\label{tab:cl-gap-analysis}

\vspace{0.5em}
\begin{minipage}[t]{0.50\textwidth}
\centering
\textbf{(a) E2E policy performance on \emph{Navhard-Base}.}\\[0.3em]
\footnotesize
\setlength{\tabcolsep}{4pt}
\begin{tabular}{l l c c c}
\toprule
\textbf{Policy} & \textbf{Obs.} & \textbf{DS} & \textbf{EPDMS} & \textbf{RC} \\
\midrule
\multirow{3}{*}{DrivoR}
  & Sim   & 42.32 & 66.99 & 62.39 \\
  & DA    & 38.92 & 66.16 & 57.74 \\
  & Ours  & \textbf{46.21} & \textbf{68.64} & \textbf{66.28} \\
\midrule
\multirow{3}{*}{DiffusionDrive}
  & Sim   & 46.19 & 61.56 & 72.67 \\
  & DA    & 32.02 & 56.39 & 55.56 \\
  & Ours  & \textbf{59.68} & \textbf{72.98} & \textbf{81.37} \\
\midrule
\multirow{3}{*}{DiffusionDriveV2}
  & Sim   & 45.87 & 57.60 & 77.67 \\
  & DA    & 21.48 & 53.06 & 38.54 \\
  & Ours  & \textbf{58.35} & \textbf{67.52} & \textbf{85.48} \\
\midrule
\multirow{3}{*}{LTF}
  & Sim   & 40.64 & 58.60 & 68.26 \\
  & DA    & 38.27 & 60.25 & 63.84 \\
  & Ours  & \textbf{53.28} & \textbf{67.84} & \textbf{77.93} \\
\bottomrule
\end{tabular}
\end{minipage}%
\hfill
\begin{minipage}[t]{0.50\textwidth}
\centering
\textbf{(b) Scorer bias \& proposal coverage failure.}\\[0.3em]
\footnotesize
\setlength{\tabcolsep}{4pt}
\begin{tabular}{l c c c}
\toprule
 & \multicolumn{2}{c}{\textbf{Scorer Bias}} & \textbf{Coverage} \\
\cmidrule(lr){2-3}
\textbf{Policy}              & \textbf{Base} & \textbf{Adv} & \textbf{Gap} \\
\midrule
DiffusionDrive              & $+3.64$  & $+8.98$  & $-12.98$ \\
DrivoR          & $+14.17$ & $+7.19$  & $-16.66$ \\
LTF$^{\dagger}$ & ---      & ---      & $-14.61$ \\
\bottomrule
\end{tabular}

\vspace{0.15em}

\textbf{(c) DrivoR on a high driving score subset of \emph{Navhard-AdvWeather}.}\\[0.3em]
\footnotesize
\setlength{\tabcolsep}{6pt}
\begin{tabular}{l c c c}
\toprule
\textbf{Condition} & \textbf{DS} & \textbf{EPDMS} & \textbf{RC} \\
\midrule
Original & \textbf{88.44} & \textbf{91.86} & \textbf{96.38} \\
Rain         & 81.70 & 86.30 & 94.10 \\
Snow         & 81.96 & 86.25 & 94.50 \\
Night        & 85.92 & 90.57 & 94.92 \\
\bottomrule
\end{tabular}
\end{minipage}
\end{table}

%% file: tex/tables/ablation.tex
\begin{table}[t]
\centering
\begin{minipage}[t]{0.56\textwidth}
\centering
\captionsetup{width=\linewidth}
\captionof{table}{Ablation on guidance. Our guidance improves layout fidelity over standard CFG.}
\label{tab:ablation-guidance}
\small
\setlength{\tabcolsep}{3pt}
\resizebox{\linewidth}{!}{%
\begin{tabular}{l c c c c c c}
\toprule
Variant & DrivoR & DD & LTF & RAP & SDv2 & \metric{}$\downarrow$ \\
\midrule
w/o TGD           & 11.25 & 22.48 & 6.92 & 10.87 &  9.38 & 12.18 \\
w/ TGD            &  \textbf{8.53} & \textbf{18.92} & \textbf{3.56} & \textbf{9.65} & \textbf{6.86} & \textbf{9.50} \\
\bottomrule
\end{tabular}%
}
\end{minipage}
\hfill
\begin{minipage}[t]{0.41\textwidth}
\centering
\captionsetup{width=\linewidth}
\captionof{table}{Ablation on diverse-scene distillation, reporting text-image CLIP score.}
\label{tab:ablation-diverse-scene}
\small
\setlength{\tabcolsep}{5pt}
\resizebox{\linewidth}{!}{%
\begin{tabular}{l c c c}
\toprule
Variant & Night & Snow & Rain \\
\midrule
w/o DSD & 0.2679 & 0.2516 & 0.2421 \\
w/ DSD  & \textbf{0.2796} & \textbf{0.2722} & \textbf{0.2475} \\
\bottomrule
\end{tabular}%
}
\end{minipage}
\end{table}

%% file: tex/main/05_limitations.tex
\section{Limitations}
\label{sec:main-limitations}

\pipeline bridges the visual gap between simulator and real-world camera observations, but our closed-loop evaluation still runs in simulation instead of on-road real-world evaluation or hardware-in-the-loop deployment given the cost and safety. Besides, the autoregressive video model also accumulates drift over extensive long-horizon rollouts, which we leave as future work.

%% file: tex/main/06_conclusion.tex
\section{Conclusion}
\label{sec:main-conclusion}

In this paper, we presented \pipeline, a generative closed-loop simulator pairing a physics simulator with an autoregressive video model; \metric, a policy-oriented Fr\'echet distance that better measures visual alignment, on which \pipeline improves over the best prior closed-loop simulators by $1.6\times$ on nuScenes and $4.7\times$ on NAVSIM; and \benchmark, a closed-loop benchmark built on \pipeline with behavior and visual variations. Our closed-loop benchmark reveals performance gaps and failure modes that prior closed-loop benchmarks miss, and points scorer calibration and proposal coverage as concrete directions for training stronger closed-loop end-to-end driving policies.

%% file: tex/supp/appendix.tex
\section{More Related Work}

\textbf{Driving Simulators and Closed-loop Benchmarks}.
Closed-loop (CL) simulators for E2E driving differ mainly in how they render the observations for policy. Game-engine simulators~\cite{dosovitskiy2017carla,gtav,muller2018sim4cv,shah2018airsim,deepdrive,kothari2021drivergym,li2022metadrive,kazemkhani2024gpudrive,gulino2024waymax} and the CL benchmarks built on them~\cite{jia2024bench,gerstenecker2026fail2drive,zhao2026bridgesim} offer full control over layout and reactive, adversarial behavior, but render using hand-built 3D assets and thus exhibit a sim-to-real visual gap that can corrupt a vision-based policy's perception.
Two recent lines of work narrow this gap, but each exhibits its own limitations. Reconstruction-based renderers~\cite{yang2024emernerf,zhou2025hugsim,gao2025rad,navsimv2,ni2025recondreamer,ni2025recondreamerrl,zhao2025drivedreamer4d} lift captured driving logs into 3D for photo-realistic rollout, but are limited to pre-captured scenarios with fixed appearance, and exhibit rendering artifacts under off-trajectory
viewpoints and dynamic-object insertion.
Generative re-rendering of simulator state~\cite{yang2025drivearena,mei2024dreamforge,zhou2024simgen,mo2025dreamland,gao2024magbeic_drive_v1,wen2024panacea,zheng2024genad,gao2024vista,hu2023gaia,lu2024infinicube} produces photo-realistic, controllable rollouts but is typically trained on small in-domain datasets such as nuScenes~\cite{caesar2020nuscenes}, inheriting a narrow visual style with weak 3D grounding. \pipeline is a hybrid approach that keeps a physics simulator for grounded, controllable, reactive state and distills an autoregressive video model from a large pretrained video model to render diverse, photo-realistic observations. It uniquely combines photorealism and visual diversity with grounded, reactive CL simulation, as shown in~\cref{tab:sim-comparison}.

\input{tex/tables/sim_comparison}

\section{Methodology Details}
We provide the training details of our autoregressive video model in this section.
Sec.~\ref{sec:supp-distillation-details} presents the three-stage distillation pipeline and its training objectives.
Sec.~\ref{sec:supp-implementation-details} reports the training hyperparameters and related implementation settings for each stage.

\subsection{Autoregressive Video Model Distillation Details}
\label{sec:supp-distillation-details}
\smallskip
\noindent
\textbf{Stage-1: Bidirectional Model}. The goal of this stage is to train a stronger teacher model that can follow the scene layout to generate coherent and continuous video frames. Following the latent video diffusion framework, we operate on VAE latents $\mathbf{x}_0 \in \mathbb{R}^{F \times H \times W \times C}$, where $F$ is the number of latent frames, $H$ is the height, $W$ is the width, and $C$ is the number of channels. Building upon the flow-matching framework, we sample Gaussian noise $\epsilon \sim \mathcal{N}(0, \mathbf{I})$ and form noised latents $\mathbf{x}_t$ by interpolating between $\mathbf{x}_0$ and $\epsilon$ over flow time $t \in [0,1]$. We train a bidirectional generator $G_{\mathrm{T}}$ with the following loss:
\begin{equation}
    \mathcal{L}_{\mathrm{flow}} = \mathbb{E}
\left[ \left\|  G_{\mathrm{T}}(\mathbf{x}_t, c, t) - (\epsilon - \mathbf{x}_0)\right\|_2^2 \right],
\end{equation}
where the condition $c = (\mathbf{c}, \text{CLIP}(\mathbf{o}_0))$ combines the scene layout $\mathbf{c}$ with the CLIP embedding of the initial camera frame $\mathbf{o}_0$, and $t$ is the flow-matching time step. After training, this model can follow the conditions and iteratively denoise random noise into visually realistic video frames.

\smallskip
\noindent
\textbf{Stage-2: Causal ODE init}. The goal of this stage is to distill the Stage-1 bidirectional teacher $G_{\mathrm{T}}$ into a few-step causal student $G_\theta$ that supports chunk-level autoregressive prediction: each chunk contains one or more latent frames with bidirectional attention inside the chunk and causal attention across chunks~\cite{huang2026self}.
Under the diffusion forcing layout, each frame is denoised with an independently noised causal context while the current frame stays noisy.
Following Self Forcing~\cite{huang2026self}, for each $\mathbf{x}_0$ we simulate the teacher reverse-time PF-ODE with $G_{\mathrm{T}}$ to obtain $\{\mathbf{x}_\tau^{\mathrm{T}}\}_{\tau \in \mathcal{T}}$, sample $t \in \mathcal{T}$ and $\mathbf{x}_t^{\mathrm{T}}$, and optimize the ODE regression objective
\begin{equation}
    \mathcal{L}_{\mathrm{ode}} = \mathbb{E}_{\mathbf{x}_0,\, t}
\left[ \left\|  G_\theta(\mathbf{x}_t^{\mathrm{T}},\, c,\, t) - \mathbf{x}_0 \right\|_2^2 \right],
\end{equation}
which distills $G_{\mathrm{T}}$ into the few-step causal student while retaining diffusion-forcing context noise during training.

\smallskip
\noindent
\textbf{Stage-3: Self Forcing}. During inference, the autoregressive model must condition each step on its own previously generated latents, but Stage-2 trains the student with ground-truth context, creating a train-test mismatch.
We close this gap with Self Forcing~\cite{huang2026self}: during training we unroll the causal student with KV caching, denoising each chunk from previously self-generated latents rather than ground truth history, and optimizing a holistic distribution-matching objective on the rollout video.
Let $\hat{\mathbf{x}}_0^{1:F} = \mathcal{R}_\theta(c, \mathcal{K})$ denote the clean latents from autoregressively rolling out $G_\theta$ under $c$.
Following the DMD generator objective in Self Forcing~\cite{huang2026self}, we define
\begin{equation}
    \mathcal{L}_{\mathrm{sf}} = \mathbb{E}_{t,\, \epsilon}
\left[ \frac{1}{2} \left\| \hat{\mathbf{x}}_0 - \mathrm{sg}\!\left( \hat{\mathbf{x}}_0 - \big( s_{\mathrm{fake}}(\hat{\mathbf{x}}_t, c, t) - s_{\mathrm{real}}(\hat{\mathbf{x}}_t, c, t) \big) \right) \right\|_2^2 \right],
\end{equation}
where $\mathrm{sg}(\cdot)$ denotes stop-gradient, $\hat{\mathbf{x}}_t = (1-t)\hat{\mathbf{x}}_0 + t\epsilon$ for $\epsilon \sim \mathcal{N}(0, \mathbf{I})$, $s_{\mathrm{real}}$ is a frozen score network initialized from $G_{\mathrm{T}}$, and $s_{\mathrm{fake}}$ is a trainable critic on the student's self-rollout.

We observe that using fully denoised latents as the KV cache leads to rapid quality drift, as high-frequency errors from the final denoising steps compound over rollout and dominate long-horizon degradation.
To mitigate this, we condition on early-timestep latents only, which empirically stabilizes rollout.

\subsection{Implementation Details}
\label{sec:supp-implementation-details}
Our training setups closely follow MotionStream~\cite{shin2025motionstream}. We use two Wan backbones~\cite{wan2025wan}: Wan~2.1 (1.3B) at $832 \times 480$ and Wan~2.2 (5B) at $1280 \times 704$; all other settings are shared. We initialize the model weights from its Wan-Fun-Control checkpoint.

In our training, Stage-1 uses batch size $64$ and learning rate $1 \times 10^{-5}$ with $10K$ steps; Stage-2 trains for $20K$ steps with batch size $64$ and learning rate $2 \times 10^{-6}$; Stage-3 trains for $1K$ steps with batch size $32$, generator learning rate $2 \times 10^{-6}$, and critic learning rate $4 \times 10^{-7}$ ($1{:}5$ updates).
For traffic-guided distillation, we use a guidance scale of $7.5$ at generation time to sample ODE trajectories and provide guidance to the student model.

\subsection{Synthetic Clip Construction for Diverse-Scene Distillation}
\label{sec:supp-synthetic-clips}

We detail the recipe for constructing the diverse synthetic clips used in Stages~2--3.

\smallskip
\noindent
\textbf{Source scenarios.} We uniformly sample 500 driving scenarios from the nuPlan-based training set (Sec.~\ref{sec:supp-training-dataset}) as layout sources. The traffic layout $\mathbf{c}$ (HD map projection and 3D bounding boxes) of each sampled clip is kept untouched, so lanes, agents, and ego motion remain aligned with the original log.

\smallskip
\noindent
\textbf{First-frame re-rendering.} We re-render only the initial frame $\mathbf{o}_0$ of each clip with the Nano-Banana~\cite{fortin2025gemini25flashimage}, prompted to change the visual appearance while preserving scene geometry. The editing prompts are sampled from the three appearance axes used in \emph{Navhard-AdvWeather}: lighting conditions, weather conditions, and road-surface conditions. We show the prompt template for changing the weather condition as follows:
\begin{quote} \small
    This is a driving-camera photo. Edit the weather and lighting only.

Do NOT move, add, remove, resize, or redraw any structure. Every lane line, every building, every billboard, every palm tree, every traffic signal, every electric pole, and every vehicle must stay in exactly the same position with exactly the same shape and the same identity. Preserve the camera angle, framing, perspective, and proportions perfectly.

Change only the season and lighting: make it a heavy snowy winter day. Cover the road shoulders, sidewalks, rooftops, billboard ledges, and palm fronds with snow. Add gentle snowflakes in the air. Replace the bright sunny sky with a flat overcast gray sky and tone down the warm sunlight to cool, diffuse winter daylight.

Output only the edited image, same resolution and same aspect ratio.
\end{quote}

\smallskip
\noindent
\textbf{Clip generation and filtering.} The Stage-1 teacher $G_{\mathrm{T}}$ rolls out a full video from the edited $\mathbf{o}_0$ and the original layout $\mathbf{c}$, producing clips with appearance follows the edited frame while ego motion and surrounding traffic follow the log. We manually review the edited samples to ensure the alignment with the appearance edit instruction, retaining $500$ synthetic scenes that are added to the Stage-2 and Stage-3 training sets.

\section{Experimental Details}

\subsection{Training Dataset}
\label{sec:supp-training-dataset}
To train and distill our autoregressive world model, we curate the dataset based on nuPlan~\cite{caesar2021nuplan}. It contains more than 20,000 driving scenario videos, each ranging from 15--20 seconds, with a total of around 90 hours of training data. To obtain the traffic-layout condition $\mathbf{c}$, we project the HDMap and 3D bounding boxes annotations into perspective view according to the camera intrinsic and extrinsic. It includes map elements like lane lines, road boundaries, crosswalks, etc., and traffic agents like vehicles and pedestrians, with each object type color-coded. The text prompt $\mathbf{p}$ for each video is generated using Qwen3-VL~\cite{Qwen3-VL} with targeted engineered prompt emphasize factual scene details like objects, geometry, motion, and context.

\subsection{Metric Details}
\pipeline adopts open-loop metrics for non-reactive open-loop simulations and closed-loop metrics for closed-loop simulations. For more motivations about metric designs, we refer the reader to~\cite{navsimv1, navsimv2, zhou2025hugsim, zhao2026bridgesim}.

\smallskip
\noindent
\emph{Open-loop metrics:} PDMS \& EPDMS. The planned trajectory output by E2E policy is scored against the logged future without being executed. The score is implemented as a weighted combination of hard safety constraints $\mathcal{C}$ and driving quality features $\mathcal{F}$, following
\begin{equation}
    \text{PDMS / EPDMS} = \Big( \prod_{sc \in \mathcal{C}} \mathbb{I}_{sc} \Big) \cdot \frac{\sum_{f \in \mathcal{F}} w_f\, v_f}{\sum_{f \in \mathcal{F}} w_f} \;\in\; [0,1],
\end{equation}
where $\mathbb{I}_{sc} \in \{0,1\}$ marks safety compliance with constraint $sc$, $v_f \in [0,1]$ is the value of quality term $f$, and $w_f$ its weight. The constraints $\mathcal{C}$ comprise NC (No At-Fault Collisions), DAC (Drivable Area Compliance), TLC (Traffic Light Compliance), and DDC (Driving Direction Compliance); the quality features $\mathcal{F}$ comprise EP (Ego Progress), LK (Lane Keeping), TTC (Time-to-Collision), C (Comfort), HC (History Comfort), and EC (Extended Comfort). For Predictive Driver Model Score (\textbf{PDMS})~\cite{navsimv1}, it uses $\mathcal{C} = \{\text{NC}, \text{DAC}, \text{DDC}\}$ and $\mathcal{F} = \{\text{EP}, \text{TTC}, \text{C}\}$. Extended Predictive Driver Model Score (\textbf{EPDMS})~\cite{navsimv2} extends it to $\mathcal{C} = \{\text{NC}, \text{DAC}, \text{DDC}, \text{TLC}\}$ and $\mathcal{F} = \{\text{EP}, \text{TTC}, \text{LK}, \text{HC}, \text{EC}\}$.

\smallskip
\noindent
\emph{Closed-loop metric: Driving Score}. For reactive closed-loop simulation, we score the full executed trajectory of the scenario by Driving Score (\textbf{DS}): global route completion (\textbf{RC}) times the mean per-frame EPDMS over the episode:
\begin{equation}
    \text{DS} = R_c \cdot \frac{1}{T} \sum_{t=1}^{T} \text{EPDMS}^{(t)},
\end{equation}
where $R_c\in[0, 1]$ represents the percentage of the route completed by the agent relative to the expert driver's path or the goal destination, $T$ is the total number of frames in the simulation episode, and $\text{EPDMS}^{(t)}$ is the EPDM score at time step $t$.

\subsection{Closed-loop Simulation Setup}

All closed-loop evaluations run with BridgeSim~\cite{zhao2026bridgesim} as the backend physics simulator, with a replan period of $\Delta_r = 5$ frames and a simulation horizon of $8$ seconds. At each iteration, the policy consumes the most recent $K = 1$ world-model-rendered frames and outputs a planned trajectory, of which the first $\Delta_r$ steps are executed.
Surrounding agents are initialized at their ground-truth pose, and their subsequent behaviors are controlled by the intelligent driver model (IDM), so they react to the ego vehicle while adhering to traffic rules.

\subsection{\benchmark Scenario Curation}
\label{sec:supp-benchmark-scenarios}

We construct \benchmark, a closed-loop benchmark to systematically diagnose the observation and behavior gaps that current E2E policies face under closed-loop deployment. It consists of a real-world log-replay base set \emph{Navhard-Base}~\cite{navsimv2} and two challenging variations: \emph{Navhard-AdvBehavior} and \emph{Navhard-AdvWeather}.

\smallskip
\noindent
\textbf{\emph{Navhard-Base}}. The base set comprises the 421 scenarios of NAVSIM \emph{navhard}~\cite{navsimv2}, each ported into the MetaDrive backend via ScenarioNet~\cite{li2023scenarionet, zhao2026bridgesim}. These scenarios cover dense urban driving with a wide range of map topologies and surrounding-agent densities, and serve as our distribution-matched reference.

\begin{wrapfigure}{r}{0.40\textwidth}
\begin{minipage}[b]{0.40\textwidth}
\centering
\includegraphics[width=\textwidth]{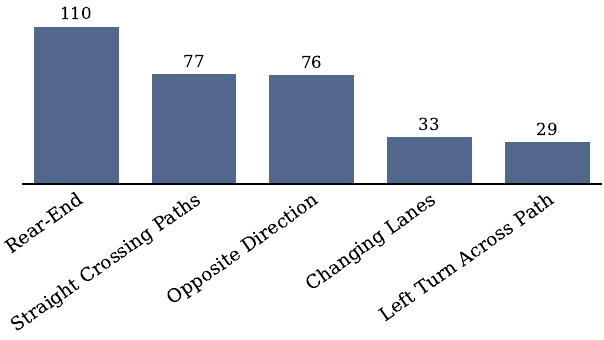}
\caption{\emph{Navhard-AdvBehavior} distribution over scenario categories.}
\label{fig:advbehavior-dist}
\end{minipage}
\end{wrapfigure}

\smallskip
\noindent
\textbf{\emph{Navhard-AdvBehavior}}. For each base scenario, we use Adv-BMT~\cite{advbmt} default setting to introduce an adversarial agent that maneuvers to provoke a collision with the ego vehicle. We review and filter to retain 325 high-quality safety-critical variants. Each variant inherits the ego route and map of its base scenario but exposes the policy to a reactive adversary, enabling a paired comparison against the base set on the same map.
These variants span five categories following the National Highway Traffic Safety Administration (NHTSA) pre-crash scenario typology~\cite{najm2007precrash,swanson2019precrash}, including \emph{Rear-End}, \emph{Straight Crossing Paths}, \emph{Opposite Direction} (head-on), \emph{Changing Lanes} (cut-in), and \emph{Left Turn Across Path}. The resulting distribution is shown in \cref{fig:advbehavior-dist}.

\smallskip
\noindent
\textbf{\emph{Navhard-AdvWeather}}. To evaluate the policy's robustness under appearance shift, we render the base scenarios under diverse conditions that vary on three axes: lighting conditions (sunrise, sunset, twilight, golden hour, blue hour, night), weather conditions (overcast, snow, rain, fog), and road-surface conditions (snow-covered, sand-covered, puddles). The text prompt and initial frame input are adapted and re-rendered correspondingly. This yields more than 5k scenario-appearance combinations for evaluating the policy.

\subsection{Closed-loop Gap Analysis}
\label{sec:supp-gap-analysis}

\benchmark exposes performance gap and failure modes that open-loop scoring and prior closed-loop benchmarks cannot reveal. On \emph{Navhard-AdvBehavior} we decompose the closed-loop performance drop to scorer bias and proposal coverage; on \emph{Navhard-AdvWeather} we measure a policy's visual robustness under appearance shift.

For the \emph{Navhard-AdvBehavior} decomposition, we keep each scoring policy's proposal set fixed and replace its learned scorer with an oracle that selects the proposal with the highest ground-truth EPDMS computed in the simulator. This enables us to measure the \emph{scorer bias gap} as the driving score decrease due to scorer mis-ranking proposals the policy generated. The \emph{coverage gap} measures the policy's proposal degradation under adversarial scenarios when scoring is already optimal.

\input{tex/tables/scorer_proposal_gap}
\input{tex/tables/weather_robustness}

\paragraph{Scorer bias widely exists across E2E policies.} \cref{tab:scorer-proposal-gap} shows that scorer bias leads to varying performance gaps, with DrivoR decreasing the most, while DiffusionDriveV2's is effectively zero. We attribute the small scorer bias of DiffusionDriveV2 to its carefully finetuned scorer using EPDMS, which effectively mitigates the scorer gap under \emph{Navhard-Base} normal scenarios. Under safety-critical distributions, this gap shifts differently for different policies. The DiffusionDrive and DiffusionDriveV2 scorer bias doubles, while DrivoR's narrows due to the Oracle performance collapses. Further analysis on DrivoR shows that on $14.1\%$ of \emph{Navhard-AdvBehavior}, its proposal set contains no candidate with positive EPDMS, versus $7.9\%$ on \emph{Navhard-Base}. \benchmark's reactive closed-loop rollouts compound the cost of scorer mis-ranking, which is different from open-loop evaluation on logged trajectories. This extends the previous open-loop scorer-mismatch analysis~\cite{ang2026clover} into the closed-loop regime, revealing the scorer bias that open-loop scoring cannot expose.

\paragraph{Proposals fail to cover feasible and recovery trajectories.} The Oracle scorer represents the upper bound that scoring can recover; the remaining performance gap is attributable to the proposal set failing to contain a feasible safe trajectory. \cref{tab:scorer-proposal-gap} shows large Oracle DS decreases from \emph{Navhard-Base} to \emph{Navhard-AdvBehavior}. Furthermore, we observe a similar scale of performance gap for a policy such as LTF that emits a single trajectory without a scorer. This verifies that this gap resides in the trajectory decoder. Analysis on DrivoR's $64$ proposal set reveals the average EPDMS of the best proposal decreases from $70.6$ to $66.7$ under safety-critical distribution. These results show the \emph{coverage gap} of proposal generation, a failure mode that only surfaces when forcing the proposal coverage outside the policy's training distribution.

\paragraph{Visual robustness is limited more by road-surface appearance than by lighting.} We evaluate DrivoR across appearance variations in \emph{Navhard-AdvWeather}, with results shown in~\cref{tab:weather-robustness}. We observe that lighting and common weather variations have little impact on the driving performance, as these conditions are also present in the training distribution. However, for conditions that obscure road elements, such as fog, snow, sand, and puddles, they result in larger performance degradation. Closed-loop safety submetrics, including NC and TTC, remain stable across conditions, suggesting that the performance drop is not driven by a safety regression. The decrease mainly comes from route completion and drivable-area compliance, indicating that the policy perceives the road semantics less reliably when its perception departs from the training distribution. This reveals a visual robustness gap related to road-surface appearance, which closed-loop benchmarks with limited visual diversity are unlikely to expose.

\section{More Visualization}

\begin{figure}[h]
\centering
\includegraphics[width=\linewidth]{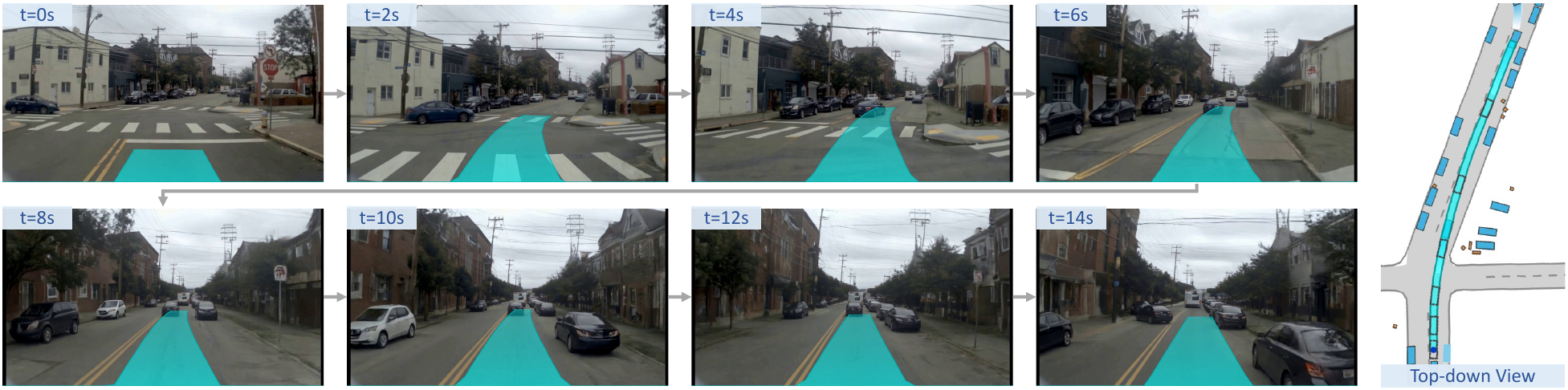}
\vspace{-1.5em}
\caption{More visualizations of closed-loop policy evaluation using \pipeline.
}
\label{fig:cl-rollout-more}
\vspace{-1.5em}
\end{figure}

We present additional qualitative examples of the closed-loop policy evaluation using DreamStream below in \cref{fig:cl-rollout-more}, demonstrating its visual realism and alignment.

%% file: tex/tables/sim_comparison.tex
\begin{table}[h]
\centering
\caption{Comparison of \pipeline with representative open-loop and closed-loop benchmarks. \emph{Photoreal.}: renders photo-realistic, low sim-to-real-gap observations. \emph{Visual Div.}: can render diverse visual appearance. \emph{Adv.}: supports reactive adversarial agents.
}
\label{tab:sim-comparison}
\small
\setlength{\tabcolsep}{5pt}
\resizebox{\textwidth}{!}{
\begin{tabular}{l|l|cccccc}
\toprule
\textbf{Name} & \textbf{Render Domain} & \textbf{Closed-loop} & \textbf{Long-horizon} & \textbf{Photoreal.} & \textbf{Visual Div.} & \textbf{Adv.} \\
\midrule
NAVSIM~\cite{navsimv1,navsimv2}            & Real-world  & $\times$     & $\times$     & \checkmark & $\times$     & $\times$     \\
Bench2Drive~\cite{jia2024bench}            & Game-engine & \checkmark & \checkmark & $\times$     & \checkmark & $\times$     \\
Fail2Drive~\cite{gerstenecker2026fail2drive} & Game-engine & \checkmark & \checkmark & $\times$     & $\times$ & \checkmark  \\
BridgeSim~\cite{zhao2026bridgesim}         & Game-engine & \checkmark & \checkmark & $\times$     & $\times$ & \checkmark  \\
HUGSIM~\cite{zhou2025hugsim}               & 3DGS Recon. & \checkmark & $\times$ & \checkmark & $\times$     & \checkmark  \\
DriveArena~\cite{yang2025drivearena}       & World Model & \checkmark & \checkmark & \checkmark & $\times$     & $\times$     \\
\midrule
\textbf{\pipeline} & World Model & \checkmark & \checkmark & \checkmark & \checkmark & \checkmark \\
\bottomrule
\end{tabular}
}
\end{table}

%% file: tex/tables/scorer_proposal_gap.tex
\begin{table}[t]
\centering
\caption{Decomposing the closed-loop performance gap on \benchmark into \emph{scorer bias} and \emph{proposal coverage}.
For each scoring policy we report driving score (DS) under its learned scorer (Learned) and an oracle scorer that selects the proposal with the highest ground-truth EPDMS (Oracle), on the base set \emph{Navhard-Base} and the adversarial set \emph{Navhard-AdvBehavior}.
\textcolor{gray}{Gray} arrows mark the two gaps: $\updownarrow$  \emph{scorer bias} gap, $\protect\longleftrightarrow$  the \emph{proposal coverage} gap. $^{\dagger}$ has no scoring head.}
\label{tab:scorer-proposal-gap}
\small
\begin{tabular}{l l c @{\hspace{5pt}} c @{\hspace{5pt}} c}
\toprule
\textbf{Policy} & \textbf{Scorer} & \shortstack{\textbf{Base}\\\textbf{DS}\,$\uparrow$} & & \shortstack{\textbf{Adv}\\\textbf{DS}\,$\uparrow$} \\
\midrule
\multirow{3}{*}{DiffusionDrive~\cite{diffusiondrive}}
  & Learned & 58.76 &  & 40.44 \\
  &         & \textcolor{gray}{$\updownarrow3.64$} &  & \textcolor{gray}{$\updownarrow8.98$} \\
  & Oracle  & 62.40 & \textcolor{gray}{$\overset{12.98}{\longleftrightarrow}$} & 49.42 \\
\midrule
\multirow{3}{*}{DiffusionDriveV2~\cite{diffusiondrivev2}}
  & Learned & 58.06 &  & 38.34 \\
  &         & \textcolor{gray}{$\updownarrow{-}0.99$} &  & \textcolor{gray}{$\updownarrow3.13$} \\
  & Oracle  & 57.07 & \textcolor{gray}{$\overset{15.60}{\longleftrightarrow}$} & 41.47 \\
\midrule
\multirow{3}{*}{DrivoR~\cite{DrivoR}}
  & Learned & 46.04 &  & 36.36 \\
  &         & \textcolor{gray}{$\updownarrow14.17$} &  & \textcolor{gray}{$\updownarrow7.19$} \\
  & Oracle  & 60.21 & \textcolor{gray}{$\overset{16.66}{\longleftrightarrow}$} & 43.55 \\
\midrule
LTF$^{\dagger}$~\cite{transfuser}
  & ---     & 51.97 & \textcolor{gray}{$\overset{14.61}{\longleftrightarrow}$} & 37.36 \\
\bottomrule
\end{tabular}%
\end{table}

%% file: tex/tables/weather_robustness.tex
\begin{table}[t]
\centering
\caption{Closed-loop evaluation of DrivoR across appearance variations in \emph{Navhard-AdvWeather}, comparing against the base scenarios. \textcolor{gray}{Gray} is the DS gap versus base.
The degradation mainly resides in route completion (RC) and drivable-area compliance (DAC) under road-surface-obscuring conditions.
}
\label{tab:weather-robustness}
\small
\setlength{\tabcolsep}{7pt}
\begin{tabular}{l|c|cc|c c c c c}
\toprule
\textbf{Condition} & \textbf{DS} & \textbf{EPDMS} & \textbf{RC} & \textbf{NC} & \textbf{DAC} & \textbf{TTC} & \textbf{LK} & \textbf{EC} \\
\midrule
Base & 47.06 & 69.73 & 66.38 & 95.90 & 85.37 & 87.07 & 95.11 & 59.87 \\
\midrule
\multicolumn{9}{l}{\emph{Lighting}} \\
Blue hour    & 47.49~\textcolor{gray}{(+0.43)} & 69.82 & 66.74 & 96.14 & 84.08 & 88.47 & 95.37 & 61.10 \\
Night        & 46.79~\textcolor{gray}{($-$0.27)} & 70.08 & 65.10 & 96.16 & 83.44 & 89.74 & 95.98 & 65.22 \\
Golden hour  & 46.64~\textcolor{gray}{($-$0.42)} & 69.70 & 65.53 & 95.98 & 84.26 & 87.94 & 95.51 & 61.25 \\
Twilight     & 46.47~\textcolor{gray}{($-$0.59)} & 70.38 & 64.32 & 96.38 & 84.39 & 89.44 & 95.24 & 61.03 \\
Sunrise      & 46.29~\textcolor{gray}{($-$0.77)} & 69.48 & 65.12 & 95.80 & 84.55 & 87.49 & 95.14 & 59.70 \\
Sunset       & 46.28~\textcolor{gray}{($-$0.78)} & 70.06 & 64.17 & 96.09 & 84.19 & 89.23 & 95.66 & 61.68 \\
\midrule
\multicolumn{9}{l}{\emph{Weather}} \\
Rain         & 47.21~\textcolor{gray}{(+0.15)} & 69.08 & 66.78 & 96.38 & 83.03 & 89.15 & 95.80 & 61.45 \\
Overcast     & 47.13~\textcolor{gray}{(+0.07)} & 69.13 & 66.75 & 96.26 & 84.06 & 86.49 & 94.92 & 59.06 \\
Snow         & 44.70~\textcolor{gray}{($-$2.36)} & 67.71 & 64.89 & 96.22 & 83.01 & 86.61 & 95.24 & 57.91 \\
Fog          & 44.50~\textcolor{gray}{($-$2.56)} & 69.73 & 62.57 & 96.18 & 84.17 & 87.40 & 95.59 & 62.21 \\
\midrule
\multicolumn{9}{l}{\emph{Road surface}} \\
Puddles      & 45.75~\textcolor{gray}{($-$1.31)} & 67.96 & 65.88 & 96.09 & 81.95 & 88.31 & 95.17 & 60.33 \\
Sand-covered & 43.42~\textcolor{gray}{($-$3.64)} & 68.47 & 62.03 & 95.88 & 82.53 & 89.03 & 95.63 & 63.38 \\
Snow-covered & 42.54~\textcolor{gray}{($-$4.52)} & 66.52 & 62.76 & 95.76 & 79.37 & 91.92 & 95.94 & 64.58 \\
\bottomrule
\end{tabular}
\end{table}